\documentclass{article}
\usepackage{spconf,amsmath,graphicx} 
\usepackage{booktabs}
\usepackage{multirow}
\usepackage{url}
\usepackage[hidelinks]{hyperref}
\newcommand*{\Scale}[2][4]{\scalebox{#1}{$#2$}}%

\newcommand\blfootnote[1]{%
  \begingroup
  \renewcommand\thefootnote{}\footnote{#1}%
  \addtocounter{footnote}{-1}%
  \endgroup
}

\title{Residual Enhanced Multi-Hypergraph Neural Network}
\name{ Jing Huang, Xiaolin Huang and Jie Yang }
\address{Department of Automation, Shanghai Jiao Tong University, Shanghai, P.R. China.}

\begin{document}
%\ninept
%
\maketitle

\blfootnote{Corresponding author: Jie Yang. This research is partly supported by NSFC,  China (No: 61876107, U1803261, 61977046), National Key R\&D Program of China (No. 2019YFB1311503), Committee of Science and Technology, Shanghai, China (No. 19510711200)}

\begin{abstract}
  Hypergraphs are a generalized data structure of graphs to model higher-order correlations among entities, which have been successfully adopted into various research domains. 
  Meanwhile, HyperGraph Neural Network (HGNN) is currently the de-facto method for hypergraph representation learning.
  However, HGNN aims at single hypergraph learning and uses a pre-concatenation approach 
  when confronting multi-modal datasets, which leads to sub-optimal exploitation of the inter-correlations of multi-modal hypergraphs.
  HGNN also suffers the over-smoothing issue, that is, its performance drops significantly when layers are stacked up.
  To resolve these issues, we propose the Residual enhanced Multi-Hypergraph Neural Network, 
  which can not only fuse multi-modal information from each hypergraph effectively, but also circumvent the over-smoothing issue associated with HGNN.
  We conduct experiments on two 3D benchmarks, the NTU and the ModelNet40 datasets, and compare against multiple  state-of-the-art methods.
  Experimental results demonstrate that both the residual hypergraph convolutions and the multi-fusion architecture can improve the performance of the base model and the combined model achieves a new state-of-the-art. Code is available at \url{https://github.com/OneForward/ResMHGNN}.

\end{abstract}
\section{Introduction}
\label{sec:intro}
% Graphs are the most ubiquitous data structure for representing pairwise relationships between entities.
% However, in real-world datasets, correlations among entities are more complex and  higher-order, for which hypergraphs are introduced.
The hypergraph structure consists of a set of vertices and hyperedges, in which a hyperedge can contain any number of vertices.
In recent years, hypergraphs have attracted increasing attention in various computer vision tasks including 3D object classification \cite{pami20Hyper}, 
3D pose estimation \cite{ijcai20:Hyper3DPose}, person re-identification \cite{HyperPersonReID}, cross-modal retrieval \cite{HyperModalRetrieval}, hypergraph-based image processing \cite{HyperIP} and video segmentation \cite{HyperVideoSeg}. 

In the meantime, multi-modal datasets are becoming more and more common since the increasing availability of multiple datasets acquired from different sources while describing the same category. When dataset for each modal can be represented by a hypergraph, 
% like visual connections, text connections and social connections on social media data, 
this leads to the multi-hypergraph learning problem.  Traditional multi-hypergraph learning models, such as tMHL \cite{pami20Hyper} and iMHL \cite{pami20Hyper, iMHL}, 
utilize the hypergraph Laplacian to design  the optimization schemes, which are shallow and require high computational cost. 

HyperGraph Neural Network (HGNN) \cite{GVCNN} is currently the de-facto method for hypergraph representation learning, after which the expressive embeddings are employed into diverse downstream tasks. 
Despite the success of HGNN, it fails to handle multi-modal dataset directly. HGNN aims at single hypergraph learning and uses a pre-concatenation approach when confronting multi-modal datasets, which leads to information loss since the inter-correlations between multi-modal hypergraphs are ignored. 

HGNN also suffers the over-smoothing issue, a phenomenon that the model performance drops significantly as the number of layers increase \cite{Chen2020:GCNII}. This degradation of learning limits HGNN to be a 2-layer model, of which the maximal exploitation of hypergraph structures can not be obtained.

In this paper, we propose the Residual enhanced Multi-HyperGraph Neural Network (ResMultiHGNN) to address the above issues.
% We conduct experiments on two real-world datasets, the NTU and the ModelNet40 datasets, for view-based 3D object classification. 
We summarize our contributions as follows:
(a) We present the first Multi-Hypergraph Neural Network for direct and parallel learning in multi-modal datasets. 
(b) We enhance the traditional hypergraph convolutions with residual connection and input mapping, which eventually help to circumvent the over-smoothing issue in deep hypergraph models.
(c) With the combined techniques, we present the first deep multi-hypergraph neural network for deep multi-modal learning. In the view-based 3D object classification task,  our model achieves a new state-of-the-art.
% Our contributions are summarized as follows:
% \begin{enumerate}
%   \item We propose the first Multi-Hypergraph Neural Network for direct and parallel learning in multi-modal datasets. 
%   \item We enhance the traditional hypergraph convolutions with residual connection and input mapping, which eventually help to circumvent the over-smoothing issue in deep hypergraph models.
%   \item With the combined techniques, we present the first deep multi-hypergraph neural network for deep multi-modal learning. In the view-based 3D object classification task,  our model achieves the state-of-the-art on ModelNet40 dataset.
% \end{enumerate}

\begin{figure*}[t]
  \centering
  \begin{picture}(500, 180)
    \put(0,0){
      \begin{minipage}[b]{\linewidth}
        \centering
        \centerline{\includegraphics[trim={0 0 0 0.5cm},clip,width=\textwidth]{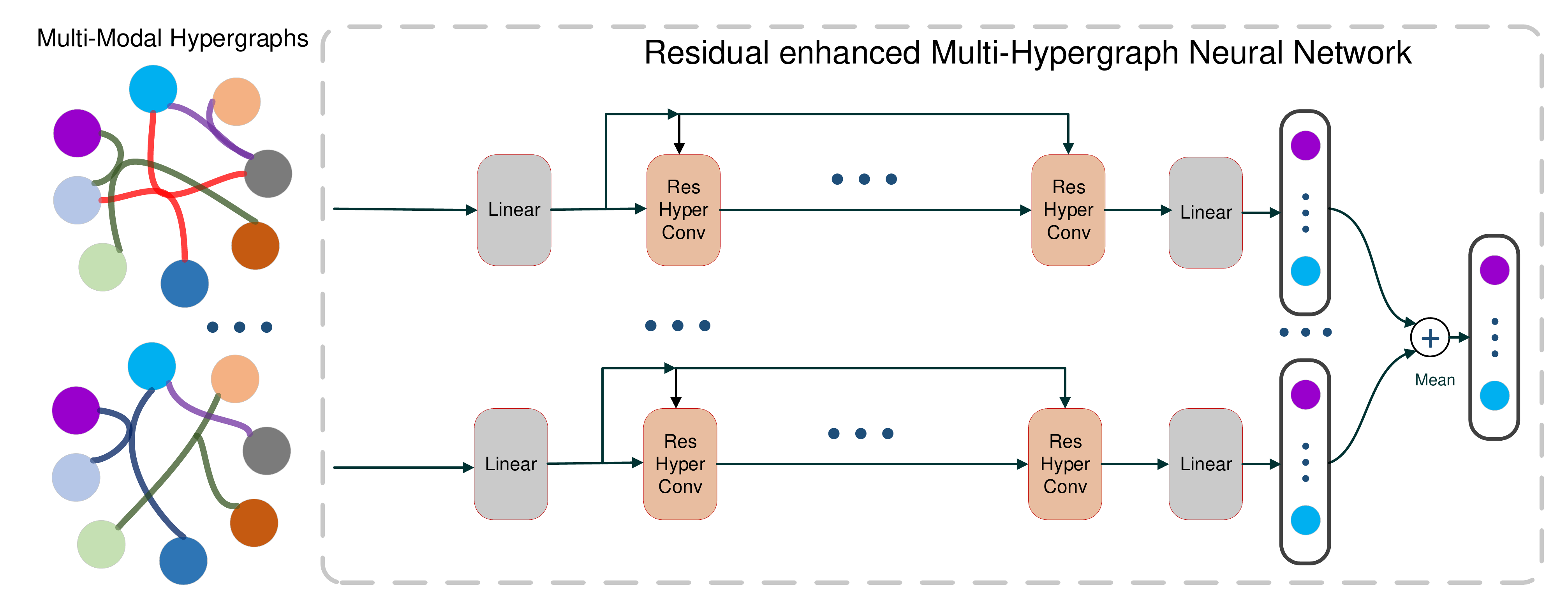}}
        
      %  \vspace{2.0cm}
        % \centerline{(a) Result 1}\medskip
      \end{minipage}
    }
    \put(110,135){$\Scale[.9]{(\mathbf{X}_1, \tilde{\mathbf{H}}_1)}$}
    \put(110,50){$\Scale[.9]{(\mathbf{X}_M, \tilde{\mathbf{H}}_M)}$}
    % \put(110,50){$(\mathbf{X}_m, \tilde{\mathbf{H}}_m)$}
  \end{picture}

  \caption{An illustration of Residual enhanced Multi-HyperGraph Neural Network.}
  \label{Fig:ResMHGNN}
\end{figure*}

\section{Background}

\subsection{Hypergraph}
Let the triplet $\mathcal{G}=(V, E, X)$ denote a hypergraph, where $V$ is a set of vertices, each hyperedge  $e \in E$ is a non-empty subset of $V$ and $\mathbf{X} \in \mathbf{R}^{|V| \times d}$ is the features of vertices. Each row of $\mathbf{X}$ denotes a $d$-dimensional feature of a vertex. 
The degree of a hyperedge $e$ is $|e|$ while the degree of a vertex $v \in V$ is defined as the number of hyperedges containing $v$.
Let $\mathbf{D}_v$ and $\mathbf{D}_e$ denote the diagonal matrices of the vertex degrees and hyperedge degrees respectively.
The hypergraph can also be characterized by $(\mathbf{X}, \mathbf{H})$, where $\mathbf{H}$ is the $|V| \times |E|$ incidence matrix with each nonzero entry $h(v, e)$ denoting $v \in e$. 
% The degree of a vertex $v \in V$ is defined as $d(v)= |\{ e \in E \}_{v \in e}|$

\begin{table}[]
  \begin{tabular}{@{}cccccc@{}}
    \toprule
  Dataset    & $N$     & $d_{\text{MVCNN}}$ & $d_{\text{GVCNN}}$ & $C$  & $\eta$  \\ \midrule
  ModelNet40 & 12311 & 4096         & 2048         & 40 & 80\% \\
  NTU        & 2012  & 4096         & 2048         & 67 & 81\% \\ \bottomrule
  \end{tabular}
  \caption{Details of the ModelNet40 and the NTU dataset. $N$ is the number of vertices. $d_{\text{MVCNN}}$ and $d_{\text{GVCNN}}$ are the dimensions of MVCNN features and GVCNN features respectively. $C$ is the number of classes and $\eta$ is the label rate.}
  \label{Tab:Dataset}
  \end{table}

\subsection{Hypergraph Neural Network}
% The hypergraph Laplacian $\tilde{\mathbf{H}}$ is defined as 
% \begin{equation}
%   \tilde{\mathbf{H}} = \mathbf{D}_v ^{-1/2} \mathbf{H} \mathbf{D}_e^{-1} \mathbf{H}' \mathbf{D}_v ^{-1/2}.
% \end{equation}
% HGNN utilizes the hypergraph Laplacian to design hypergraph convolution.
% Let $\mathbf{W}^{(l)}$ and $\mathbf{X}^{(l)}$ denote the learnable parameters and the input features in the $l$-th layer respectively, then the update rule of HGNN is formulated as 
% \begin{equation}
%   \mathbf{X}^{(l+1)} = \sigma \left( \tilde{\mathbf{H}} \mathbf{X}^{(l)} \mathbf{W}^{(l)} \right),
% \end{equation}
% where $\sigma$ is the activation function.

% The hypergraph Laplacian
HGNN utilizes the hypergraph Laplacian $\tilde{\mathbf{H}}$ to design hypergraph convolution, where  $\tilde{\mathbf{H}}$ is defined as 
\begin{equation}
  \tilde{\mathbf{H}} = \mathbf{D}_v ^{-1/2} \mathbf{H} \mathbf{D}_e^{-1} \mathbf{H}' \mathbf{D}_v ^{-1/2}.
\end{equation}
Let $\mathbf{W}^{(l)}$ and $\mathbf{X}^{(l)}$ denote the learnable parameters and the input features in the $l$-th layer respectively, then the message passing process of HGNN is formulated as 
\begin{equation}
  \mathbf{X}^{(l+1)} = \sigma \left( \tilde{\mathbf{H}} \mathbf{X}^{(l)} \mathbf{W}^{(l)} \right),
\end{equation}
where $\sigma$ is the activation function.

% A vanilla HGNN contains only 2 layers.
% A vanilla HGNN contains only 2 layers and in node classification task, additional softmax layer is added after the final embeddings for predicting the probabilities of node labels.

\begin{table*}[htb]
  \small
  \centering
  \begin{tabular}{@{}cc|rrrrrr|rrrrrr@{}}
  \toprule
  \multirow{2}{*}{Dataset}    & \multicolumn{1}{c|}{\multirow{2}{*}{Method/Layers}} & \multicolumn{6}{c}{Full}                                                  & \multicolumn{6}{|c}{Balanced}                                     \\  
                              & \multicolumn{1}{c|}{}                         & \multicolumn{1}{c}{2} & \multicolumn{1}{c}{4} & \multicolumn{1}{c}{8} & \multicolumn{1}{c}{16} & \multicolumn{1}{c}{32} & \multicolumn{1}{c}{64}  & \multicolumn{1}{|c}{2} & \multicolumn{1}{c}{4} & \multicolumn{1}{c}{8} & \multicolumn{1}{c}{16} & \multicolumn{1}{c}{32} & \multicolumn{1}{c}{64}  \\ \midrule
  \multirow{4}{*}{ModelNet40} & HGNN                                        & \textbf{96.88} & 96.68          & 96.43          & 79.58          & 4.34  & 4.13  & \textbf{97.79} & 97.65 & 97.55          & 88.27 & 5.65  & 4.15           \\
                              & MultiHGNN                                   & \textbf{97.45} & 97.16          & 96.56          & 86.31          & 4.05  & 4.05  & \textbf{98.49} & 98.29 & 97.49          & 93.51 & 4.15  & 2.77           \\
                              & ResHGNN                                     & 97.49          & 97.49          & \textbf{97.53} & 97.49          & 97.49 & 97.41 & 98.14          & 98.14 & 98.19          & 98.19 & 98.16 & \textbf{98.29} \\
                              & ResMultiHGNN                                & \textbf{98.02} & 97.81          & 97.93          & 97.97          & 97.85 & 97.69 & 98.67          & 98.66 & \textbf{98.74} & 98.68 & 98.64 & 98.71          \\ \midrule
  \multirow{4}{*}{NTU2012}    & HGNN                                        & \textbf{83.65} & 82.57          & 81.50          & 55.50          & 5.09  & 5.09  & \textbf{90.31} & 89.63 & 87.67          & 41.87 & 2.30  & 6.10           \\
                              & MultiHGNN                                   & \textbf{85.26} & 84.18          & 82.31          & 70.78          & 6.70  & 4.56  & \textbf{90.58} & 90.52 & 88.55          & 77.98 & 20.66 & 3.18           \\
                              & ResHGNN                                     & 84.99          & \textbf{85.52} & \textbf{85.52} & 85.26          & 85.26 & 85.26 & \textbf{91.80} & 91.33 & 91.26          & 91.40 & 91.67 & 91.46          \\
                              & ResMultiHGNN                                & 85.79          & \textbf{86.86} & 85.79          & 86.06          & 85.26 & 85.26 & 91.94          & 91.80 & \textbf{92.14} & 91.73 & 91.80 & 92.01 \\ \bottomrule
  \end{tabular}
      \caption{Summaries of classification accuracy (\%) results with different depths.  \textit{Full} denotes the experiments with full training labels while \textit{Balanced} denotes the experiments with balanced subset of original training labels. The result of the best performed model for each dataset is bolded.}
    \label{Tab:layers}
  \end{table*}

    \begin{figure*}[htb]
      \centering
      \begin{minipage}[b]{.48\linewidth}
        \centering
        \centerline{\includegraphics{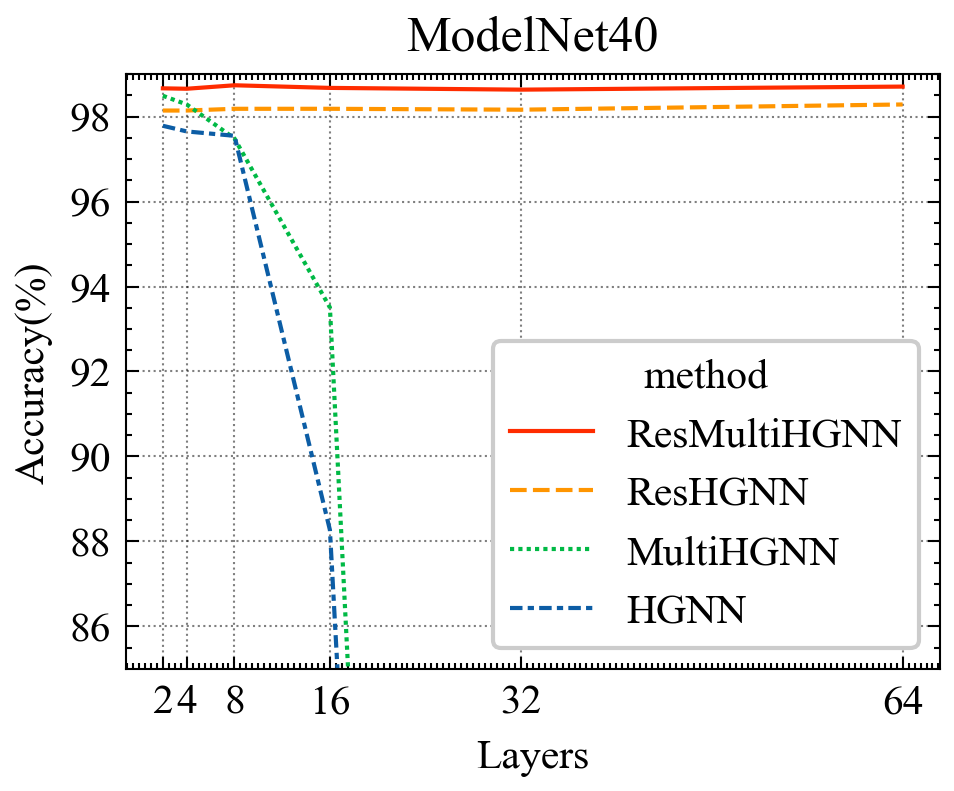}}
      %  \vspace{2.0cm}
        % \centerline{(a) Result 1}\medskip
      \end{minipage}
      \hfill
      \begin{minipage}[b]{.48\linewidth}
        \centering
        \centerline{\includegraphics{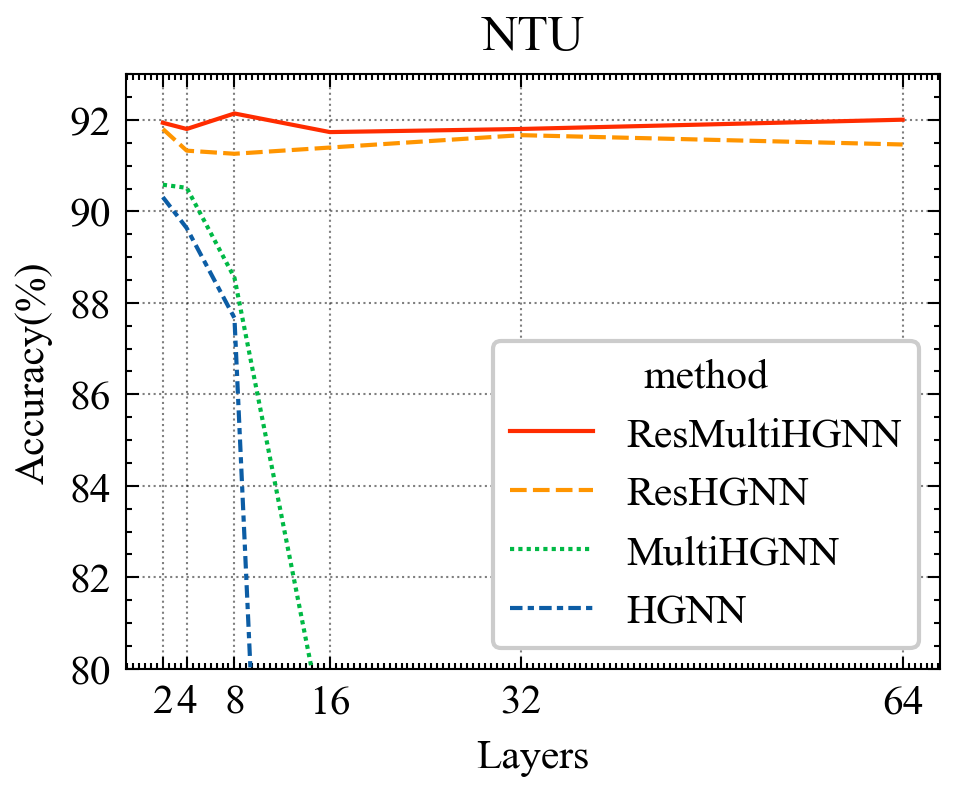}}
      %  \vspace{1.5cm}
        % \centerline{(b) Results 3}\medskip
      \end{minipage}
      \caption{The performance of different hypergraph neural networks v.s. layers in \textit{3D Object Classification Task (Balanced)}.}
      \label{Fig:layer}
    \end{figure*}

      \begin{figure*}[htb]
        \centering
        \begin{minipage}[b]{.48\linewidth}
          \centering
          \centerline{\includegraphics{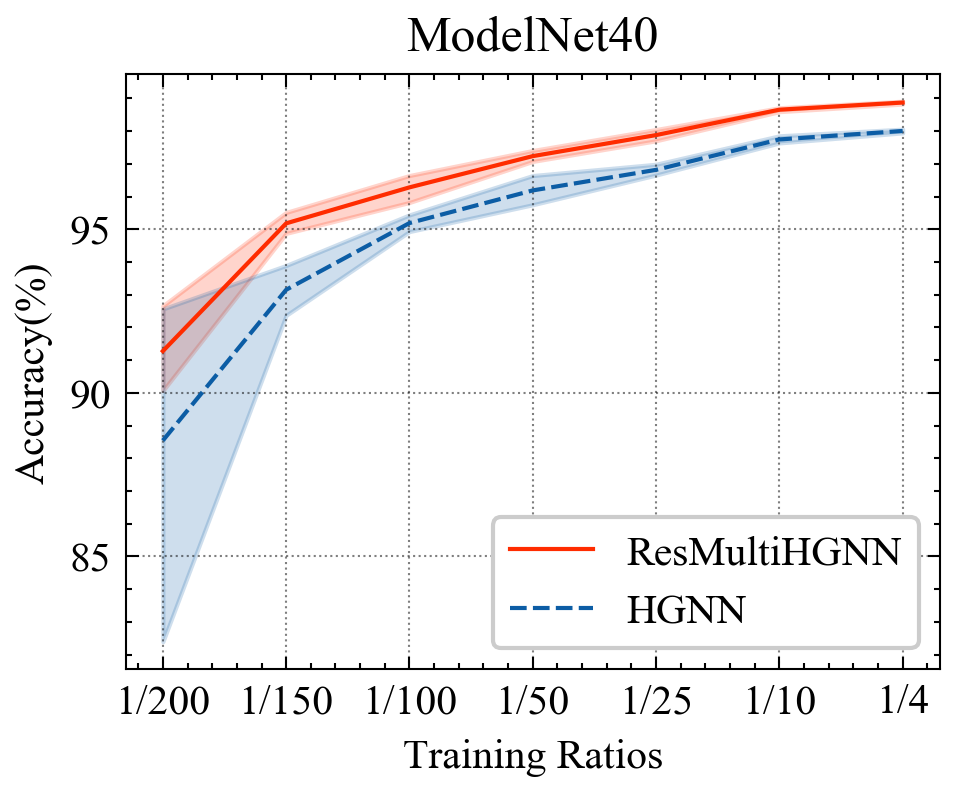}}
        %  \vspace{2.0cm}
          % \centerline{(a) Result 1}\medskip
        \end{minipage}
        \hfill
        \begin{minipage}[b]{.48\linewidth}
          \centering
          \centerline{\includegraphics{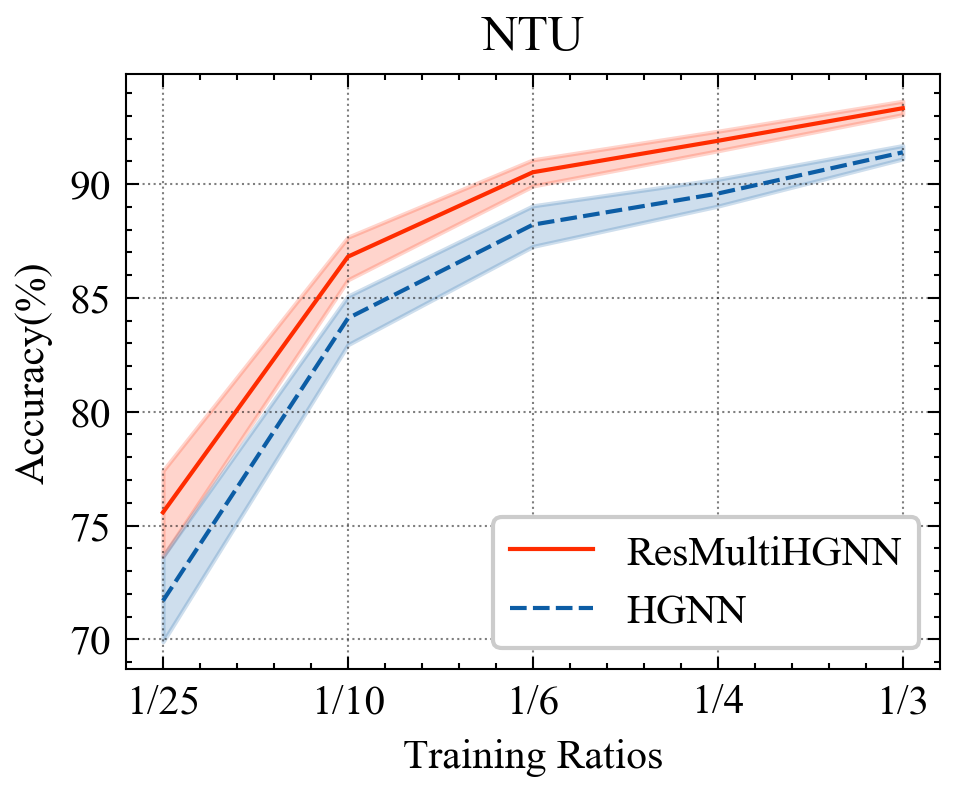}}
        %  \vspace{1.5cm}
          % \centerline{(b) Results 3}\medskip
        \end{minipage}
        \caption{Stability analysis.  The performance of HGNN and ResMultiHGNN v.s. different ratios of training labels.}
        \label{Fig:splits}
      \end{figure*}

\section{Proposed Methods}
\subsection{Multi-Hypergraph Neural Network}
Given a multi-modal dataset $\left\{ (\mathbf{X}_1, \mathbf{H}_1), \ldots, (\mathbf{X}_M, \mathbf{H}_M) \right\}$ containing $M$ hypergraphs, HGNN pre-concatenates these hypergraphs into a larger hypergraph $(\hat{\mathbf{X}}, \hat{\mathbf{H}})$ and then conducts the single hypergraph learning. 

To better exploit the inter-relations between different modals, we propose the Multi-Hypergraph Neural Network (MultiHGNN), which include $M$ distinct branches to extract high-level information from each modal in parallel. As illustrated by Fig.~\ref{Fig:ResMHGNN}, embeddings from multi-modal hypergraphs are combined together and then output for the downstream tasks. Notice that the gradients in each branch are back-propagated respectively, therefore the inner structure for each modal can be fully explored. 
In practice we use the simple \textit{Mean} function to combine the multiple embeddings. 

Formally, the output of MultiHGNN is formulated as 
\begin{equation}
  \Scale[0.85]{
    \text{MultiHGNN}\left(\left\{ (\mathbf{X}_m, \tilde{\mathbf{H}}_m) \right\}_{m=1}^M\right)= \frac{1}{M} \sum_{m=1}^{M} {\text{HGNN}(\mathbf{X}_m, \tilde{\mathbf{H}}_m) }
  },
\end{equation}
where $\tilde{\mathbf{H}}_m$ is the hypergraph Laplacian of $\mathbf{H}_m$.

\subsection{Residual Hypergraph Convolution}

In computer vision, it is well known that residual connections \cite{he2016deep:resnet} are the key components for making powerful and extremely deep networks.
Drawing inspiration from GCNII \cite{Chen2020:GCNII}, we enhance the vanilla hypergraph convolution with initial residual and identity mapping. Given the single hypergraph Laplacian $\tilde{\mathbf{H}}$, we define the propagation process of residual enhanced hypergraph convolution in the $l$-th layer as 
\begin{equation}
  \Scale[.9]
 { \mathbf{X}^{(l+1)} = \sigma \left( 
  \left((1 - \alpha_l)  \tilde{\mathbf{H}}  \mathbf{X}^{(l)} + \alpha_l \mathbf{X}^{(0)}\right) \left( (1-\beta_l) \mathbf{I} + \beta_l \mathbf{W}^{(l)} \right)
    \right)},
\end{equation}
where $\alpha_l, \beta_l$ are hyperparameters and $\mathbf{I}$ is the identity matrix.

We present the ResHGNN by naively stacking multiple blocks of  residual hypergraph convolution, which empirically circumvent the over-smoothing issue associated with HGNN.

\subsection{Deep Multi-Hypergraph Neural Network }

We combine both techniques to build very deep multi-hypergraph neural network, denoted as ResMultiHGNN. In each branch, additional Linear transforms are added in the first and last layer, then the residual hypergraph convolutions are employed to propagate information over the distinct hypergraph structure. The deep embeddings from each branch are finally combined for downstream tasks.

\section{Experiments}

\subsection{Datasets and Experimental Setup}
We evaluate the performance of the proposed method on view-based 3D object classification task. We use the Princeton ModelNet dataset \cite{ModelNet40} and National Taiwan University (NTU) 3D model dataset as testing benchmarks. 
The ModelNet40 dataset is composed of 12,311 3D CAD models from 40 popular object categories, which is the publicly used subset of Princeton ModelNet dataset \cite{ModelNet40}. We use the same training/testing splits as \cite{HGNN}, where the training split contains 9,843 objects  and the testing split contains 2,468 objects. 
The NTU dataset \cite{NTU2012} includes 2012 3D objects from 67 categories, such as \textit{boat}, 
% \textit{bomb}, \textit{book}, \textit{car}, \textit{chair}, \textit{chip}, \textit{cup}, \textit{driver}, \textit{facemask}, \textit{fence}, \textit{flashlight}, \textit{guitar}, \textit{gun}, \textit{hat}, \textit{helicopter}, \textit{hydrant}, \textit{knife}, \textit{leda}, \textit{map}, 
\textit{motorcycle}, \textit{train}, and \textit{truck}. We closely follow \cite{HGNN} and use the same training/testing splits. Details of the datasets are listed in Table~\ref{Tab:Dataset}.

For fair comparison, we employ the same shape features as \cite{HGNN}, which are extracted from two multi-view based 3D shape descriptors, i.e. Multi-View Convolutional Neural Net Network features (MVCNN) \cite{MVCNN} and Group-View Convolutional Neural Network (GVCNN) \cite{GVCNN} features. The hypergraph of each modal  is constructed based on respective features. Each hyperedge contains the 10 nearest neighbors of the central vertex/object. The multi-hypergraphs can be directly fed into the MultiHGNNs, whereas HGNNs only accept the pre-concatenated single hypergraph as input.

Note that the original training splits in both datasets contain unbalanced labels. We further conduct experiments by using only the balanced \textit{subset} of original training labels and predict the results on the \textit{rest} samples.
We implement all models with Pytorch \cite{PyTorch}.
% For all models, we set the dropout rate as 0.5, the hidden layer dimensions as 128 and use the Adam optimizer with the learning rate of 0.001. 
We release our code publicly in \href{https://github.com/OneForward/ResMHGNN}{GitHub} for reproducible experiments.
% We use the Adam optimizer with the learning rate of 0.001 in all experiments. 
% \usepackage{booktabs}

\subsection{Experimental Results}
  
% We compare our models against the following baselines,

% \begin{itemize}
%   \item 
% \end{itemize}

% \subsubsection{Comparison of Proposed Methods against HGNN}

Table~\ref{Tab:layers} reports the performance of proposed methods against HGNN with different depths, which can also be viewed as the ablation study of multi-fusion structure and residual hypergraph convolution. We visualize the results in Fig~\ref{Fig:layer} for better comparison. Based on the table and the figure, we summarize the  observations as follows: (1) Regardless of adding the residual hypergraph convolution or not, MultiHGNNs are consistently better than HGNNs,  which verifies the effectiveness of the proposed multi-fusion architecture. (2) Residual connection can consistently enhance the performance of HGNN and MultiHGNN. It is worthwhile to note that with the number of layers increase, residual enhanced models maintain stable performance whereas performances of vanilla models  deteriorate significantly. (3) ResMultiHGNN outperforms all other methods on both datasets. 
% achieves the state-of-the-art on both datasets.

% \begin{enumerate}
%   \item Multi-Hypergraph Neural Networks are consistently better than normal Hypergraph Neural Networks, regardless of adding the residual component or not, which verifies the effectiveness of the proposed multi-fusion method. 
%   \item Residual connection can consistently enhance the performance of vanilla Hypergraph or Multi-Hypergraph Neural Networks. It is worthwhile to note that with the number of layers increase, residual enhanced models maintain stable performance whereas vanilla model performances deteriorate significantly.
%   \item When both methods combined together, the residual enhanced multi-hypergraph neural network achieves the state-of-the-art on both datasets.
% \end{enumerate}

We also point out that only using the balanced \textit{subset} of training labels and testing on the \textit{rest} all samples produce much better results than using full training labels. This indicates the importance of balanced distribution of labels in training samples for hypergraph learning.

Table~\ref{Tab:ModelNet40} summarizes the classification accuracy of proposed ResMultiHGNN model against multiple recent state-of-the-art methods on ModelNet40 dataset. We see that ResMultiHGNN achieves a new state-of-the-art against the object recognition methods and the hypergraph learning methods, which demonstrates the power of multi-modal fusions and residual enhanced deep structures.
% of multi-modal fusions and deep structures.of multi-modal fusions and deep structures.of multi-modal fusions and deep structures.
% \subsubsection{Comparison with SOTAs on ModelNet40}

\subsection{Stability Analysis}

We further investigate the stability of our method against HGNN by modifying the ratios of training labels.  
We conduct all experiments with 8 different seeds and report the best performed models with optimal layers, as visualized in Fig.~\ref{Fig:splits}. We observe that ResMultiHGNN consistently shows better performances than HGNN in all training ratios with gains around 2\% and 3\%. HGNN exhibits higher variance whereas  ResMultiHGNN is slightly more stable, especially when training ratios are small. 

\begin{table}[]
  \centering
  \begin{tabular}{@{}cc@{}}
  \toprule
  Model        & Accuracy      \\ \midrule
  % PointNet \cite{PointNet}     & 89.2          \\
  PointNet++ \cite{PointNet++}   & 90.7          \\
  % PointCNN \cite{PointCNN}    & 91.8          \\
  % SO-Net \cite{SO-Net}      & 93.4          \\
  
  LP-3DCNN \cite{LP-3DCNN} & 92.1 \\ 
  RS-CNN \cite{RS-CNN} & 93.6 \\ 
  tMHL \cite{iMHL}        & 96.2          \\
  HGNN  \cite{HGNN}       & 96.9          \\
  iMHL \cite{iMHL, pami20Hyper}        & 97.2         \\
  % RotationNet \cite{RotationNet} & 97.4 \\ 
  ResMultiHGNN & \textbf{98.0} \\ \bottomrule
  \end{tabular}
  \caption{Experimental results of multiple recent state-of-the-art methods on ModelNet40 dataset.}
  \label{Tab:ModelNet40}
  \end{table}

\section{Conclusion}

In this paper, we propose the first Multi-Hypergraph Neural Network for multi-modal learning, and enhance it with the residual connections to build deep structures. Extensive experiments demonstrate the effectiveness of proposed Multi-fusion architecture and Residual Hypergraph Convolution. ResMultiHGNN enjoys both gains and shows stable and better results against HGNN.

% \section{COPYRIGHT FORMS}
% \label{sec:copyright}

% You must include your fully completed, signed IEEE copyright release form when
% form when you submit your paper. We {\bf must} have this form before your paper
% can be published in the proceedings. 

\newpage
\bibliographystyle{IEEEbib}
\bibliography{icip21}

\begin{thebibliography}{10}

\bibitem{pami20Hyper}
Yue Gao, Zizhao Zhang, Haojie Lin, Xibin Zhao, Shaoyi Du, and Changqing Zou,
\newblock ``Hypergraph learning: Methods and practices,''
\newblock {\em IEEE Transactions on Pattern Analysis and Machine Intelligence},
  2020.

\bibitem{ijcai20:Hyper3DPose}
Shengyuan Liu, Pei Lv, Yuzhen Zhang, Jie Fu, Junjin Cheng, Wanqing Li, Bing
  Zhou, and Mingliang Xu,
\newblock ``Semi-dynamic hypergraph neural network for 3d pose estimation,''
\newblock in {\em Proceedings of the Twenty-Ninth International Joint
  Conference on Artificial Intelligence, {IJCAI-20}}, Christian Bessiere, Ed.,
  2020,
\newblock Main track.

\bibitem{HyperPersonReID}
L.~{An}, X.~{Chen}, S.~{Yang}, and X.~{Li},
\newblock ``Person re-identification by multi-hypergraph fusion,''
\newblock {\em IEEE Transactions on Neural Networks and Learning Systems}, vol.
  28, no. 11, pp. 2763--2774, 2017.

\bibitem{HyperModalRetrieval}
J.~{Gao}, W.~{Zhang}, Z.~{Chen}, and F.~{Zhong},
\newblock ``Hdmfh: Hypergraph based discrete matrix factorization hashing for
  multimodal retrieval,''
\newblock in {\em ICASSP 2020 - 2020 IEEE International Conference on
  Acoustics, Speech and Signal Processing (ICASSP)}, 2020, pp. 1923--1927.

\bibitem{HyperIP}
Songyang Zhang, Shuguang Cui, and Zhi Ding,
\newblock ``Hypergraph-based image processing,''
\newblock in {\em {IEEE} International Conference on Image Processing, {ICIP}
  2020, Abu Dhabi, United Arab Emirates, October 25-28, 2020}. 2020, pp.
  216--220, {IEEE}.

\bibitem{HyperVideoSeg}
X.~{Lv}, L.~{Wang}, Q.~{Zhang}, N.~{Zheng}, and G.~{Hua},
\newblock ``Video object co-segmentation from noisy videos by a multi-level
  hypergraph model,''
\newblock in {\em 2018 25th IEEE International Conference on Image Processing
  (ICIP)}, 2018, pp. 2207--2211.

\bibitem{iMHL}
Zizhao Zhang, Haojie Lin, Xibin Zhao, Rongrong Ji, and Yue Gao,
\newblock ``Inductive multi-hypergraph learning and its application on
  view-based 3d object classification,''
\newblock {\em {IEEE} Trans. Image Process.}, vol. 27, no. 12, pp. 5957--5968,
  2018.

\bibitem{GVCNN}
Yifan Feng, Zizhao Zhang, Xibin Zhao, Rongrong Ji, and Yue Gao,
\newblock ``{GVCNN:} group-view convolutional neural networks for 3d shape
  recognition,''
\newblock in {\em 2018 {IEEE} Conference on Computer Vision and Pattern
  Recognition, {CVPR} 2018, Salt Lake City, UT, USA, June 18-22, 2018}. 2018,
  pp. 264--272, {IEEE} Computer Society.

\bibitem{Chen2020:GCNII}
Ming Chen, Zhewei Wei, Zengfeng Huang, Bolin Ding, and Yaliang Li,
\newblock ``{Simple and Deep Graph Convolutional Networks},''
\newblock {\em International Conference on Machine Learning}, 2020.

\bibitem{he2016deep:resnet}
Kaiming He, Xiangyu Zhang, Shaoqing Ren, and Jian Sun,
\newblock ``Deep residual learning for image recognition,''
\newblock in {\em Proceedings of the IEEE conference on computer vision and
  pattern recognition}, 2016, pp. 770--778.

\bibitem{ModelNet40}
Zhirong Wu, Shuran Song, Aditya Khosla, Fisher Yu, Linguang Zhang, Xiaoou Tang,
  and Jianxiong Xiao,
\newblock ``3d shapenets: {A} deep representation for volumetric shapes,''
\newblock in {\em {IEEE} Conference on Computer Vision and Pattern Recognition,
  {CVPR} 2015}, 2015, pp. 1912--1920.

\bibitem{HGNN}
Yifan Feng, Haoxuan You, Zizhao Zhang, Rongrong Ji, and Yue Gao,
\newblock ``Hypergraph neural networks,''
\newblock in {\em The Thirty-Third {AAAI} Conference on Artificial
  Intelligence, {AAAI} 2019}, 2019, pp. 3558--3565.

\bibitem{NTU2012}
Ding{-}Yun Chen, Xiao{-}Pei Tian, Yu{-}Te Shen, and Ming Ouhyoung,
\newblock ``On visual similarity based 3d model retrieval,''
\newblock {\em Comput. Graph. Forum}, vol. 22, no. 3, pp. 223--232, 2003.

\bibitem{MVCNN}
Hang Su, Subhransu Maji, Evangelos Kalogerakis, and Erik~G. Learned{-}Miller,
\newblock ``Multi-view convolutional neural networks for 3d shape
  recognition,''
\newblock in {\em 2015 {IEEE} International Conference on Computer Vision,
  {ICCV} 2015, Santiago, Chile, December 7-13, 2015}. 2015, pp. 945--953,
  {IEEE} Computer Society.

\bibitem{PyTorch}
Adam Paszke, Sam Gross, Francisco Massa, Adam Lerer, James Bradbury, Gregory
  Chanan, Trevor Killeen, Zeming Lin, Natalia Gimelshein, Luca Antiga, Alban
  Desmaison, Andreas Kopf, Edward Yang, Zachary DeVito, Martin Raison, Alykhan
  Tejani, Sasank Chilamkurthy, Benoit Steiner, Lu~Fang, Junjie Bai, and Soumith
  Chintala,
\newblock ``Pytorch: An imperative style, high-performance deep learning
  library,''
\newblock in {\em Advances in Neural Information Processing Systems},
  H.~Wallach, H.~Larochelle, A.~Beygelzimer, F.~d\textquotesingle
  Alch\'{e}-Buc, E.~Fox, and R.~Garnett, Eds. 2019.

\bibitem{PointNet++}
Charles~Ruizhongtai Qi, Li~Yi, Hao Su, and Leonidas~J. Guibas,
\newblock ``Pointnet++: Deep hierarchical feature learning on point sets in a
  metric space,''
\newblock in {\em Advances in Neural Information Processing Systems}, 2017, pp.
  5099--5108.

\bibitem{LP-3DCNN}
Sudhakar Kumawat and Shanmuganathan Raman,
\newblock ``{LP-3DCNN:} unveiling local phase in 3d convolutional neural
  networks,''
\newblock in {\em {IEEE} Conference on Computer Vision and Pattern Recognition,
  {CVPR} 2019, Long Beach, CA, USA, June 16-20, 2019}. 2019, pp. 4903--4912,
  Computer Vision Foundation / {IEEE}.

\bibitem{RS-CNN}
Yongcheng Liu, Bin Fan, Shiming Xiang, and Chunhong Pan,
\newblock ``Relation-shape convolutional neural network for point cloud
  analysis,''
\newblock in {\em {IEEE} Conference on Computer Vision and Pattern Recognition,
  {CVPR} 2019, Long Beach, CA, USA, June 16-20, 2019}. 2019, pp. 8895--8904,
  Computer Vision Foundation / {IEEE}.

\end{thebibliography}

\end{document}